\documentclass{article}
\usepackage{amssymb}
\usepackage{graphicx}
\usepackage{multirow}
\usepackage[table]{xcolor}
\usepackage{url}
\usepackage{epsfig} 
\usepackage{mathptmx} 
\usepackage{times} 
\usepackage{amsmath} 
\usepackage{soul}
\usepackage{caption}
\usepackage{xcolor}
\sethlcolor{yellow}
\usepackage{dsfont}
\usepackage{booktabs}
\usepackage{makecell}
\usepackage{wrapfig}
\usepackage{algorithm}
\usepackage{algpseudocode}
\usepackage{graphicx}
\usepackage{subcaption}
\usepackage[final]{corl_2025} 

\newcommand{\joint}{\mathbf{J}}
\newcommand{\armjoint}{\joint^{a}}
\newcommand{\handjoint}{\joint^{h}}

\newcommand{\obs}{\mathcal{O}}

\newcommand{\aset}{\mathcal{A}}

\newcommand{\E}{\mathbb{E}}

\newcommand{\policyaction}{\mathbf{a}}
\newcommand{\baseaction}{\policyaction^{a}}
\newcommand{\handaction}{\policyaction^{h}}

\title{ClutterDexGrasp: A Sim-to-Real System for General Dexterous Grasping in Cluttered Scenes}

\author{
  Zeyuan Chen$^{1,2*}$, Qiyang Yan$^{1,2*}$, Yuanpei Chen$^{1,3*}$ \\
  \textbf{Tianhao Wu$^{1,2 \dagger}$, Jiyao Zhang$^{1,2 \dagger}$, Zihan Ding$^{4}$, Jinzhou Li$^{1,2}$, Yaodong Yang$^{1,3}$, Hao Dong$^{1,2 \ddagger}$}\\
  \\
  $^{1}$CFCS, School of Computer Science, Peking University \\
  $^{2}$PKU-AgiBot Lab, $^{3}$PKU-PsiBot Lab, $^{4}$Princeton University\\
  \\
  $^*$Equal Contribution, 
  $^\dagger$Project leader, 
  $^\ddagger$Corresponding author\\
}

\begin{document}
\maketitle


\begin{figure}[h!]
    \vspace{-.3in}
    \centering
    \includegraphics[width=1\linewidth]{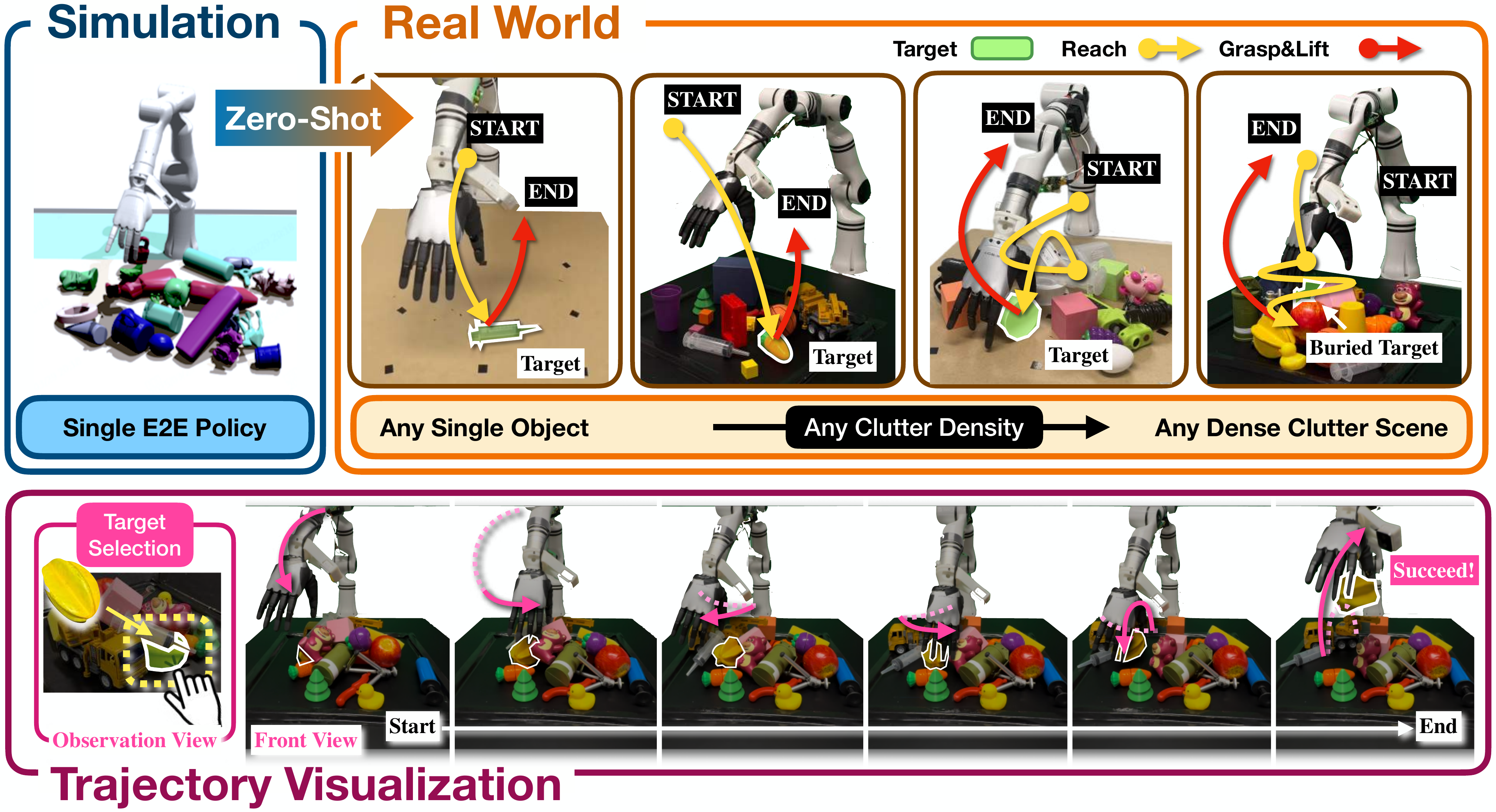}
    \caption{ClutterDexGrasp achieves zero-shot sim-to-real transfer for closed-loop target-oriented dexterous grasping in cluttered scenes, enabling robust generalization across diverse objects and cluttered scenes, even with severe object occlusion.
    }
    \label{fig:teaser}
    \vspace{-2.5mm}
\end{figure}

\begin{abstract}
    Dexterous grasping in cluttered scenes presents significant challenges due to diverse object geometries, occlusions, and potential collisions. Existing methods primarily focus on single-object grasping or grasp-pose prediction without interaction, which are insufficient for complex, cluttered scenes. Recent vision-language-action models offer a potential solution but require extensive real-world demonstrations, making them costly and difficult to scale. To address these limitations, we revisit the sim-to-real transfer pipeline and develop key techniques that enable zero-shot deployment in reality while maintaining robust generalization. We propose ClutterDexGrasp, a two-stage teacher-student framework for closed-loop target-oriented dexterous grasping in cluttered scenes. The framework features a teacher policy trained in simulation using clutter density curriculum learning, incorporating both a geometry- and spatially-embedded scene representation and a novel comprehensive safety curriculum, enabling general, dynamic, and safe grasping behaviors.
    Through imitation learning, we distill the teacher's knowledge into a student 3D diffusion policy (DP3) that operates on partial point cloud observations. To the best of our knowledge, this represents the first zero-shot sim-to-real closed-loop system for target-oriented dexterous grasping in cluttered scenes, demonstrating robust performance across diverse objects and layouts. More details and videos are available at \url{https://clutterdexgrasp.github.io/}.
\end{abstract}
\vspace{-2.5mm}

\section{Introduction}

Humans can use their hands to grasp diverse objects in cluttered scenes \citep{manipulation_behavior}, which serves as the foundation for performing subsequent manipulation tasks. Achieving such dexterity is critical for applying a dexterous hand in complex real-world environments \citep{IHM_survey, general_system_IHM}. While current methods have made significant progress in generating grasp poses for single objects~\citep{xu2024dexterous, lu2024ugg, zhong2025dexgrasp} or closed-loop grasping on single-object tabletop setups~\citep{dexpoint, wu2023learning, wan2023unidexgrasp++, huang2024efficient, lumdextrah, singh2024dextrah}, cluttered scenes present substantially different challenges. These scenes contain many objects with diverse geometries that are close together or even overlapping, leading to complex dynamics and occlusions during the grasping process \citep{fanganydexgrasp, zhang2024dexgraspnet}. 

To enable dexterous grasping in cluttered scenes, some works first generate a non-collision grasp pose dataset specifically for cluttered scenes, then train a model to generate grasp poses and execute via motion planning~\citep{wei2022dvgg, zhang2024dexgraspnet, fanganydexgrasp}. However, such methods are open-loop and cannot handle unexpected changes. To achieve a closed-loop grasping policy, current methods rely on real-world imitation learning (IL)\citep{zhong2025dexgraspvla, il_survey, il_tilde} or sim-to-real reinforcement learning (RL) \citep{deep_rl, openai_hand, deep_rl2}. However, real-world IL requires a large amount of human-collected data to generalize across diverse cluttered scenes\citep{zhong2025dexgraspvla, imitation_demo_difficult, dagger}, which is costly and lacks scalability. Current RL-based methods, which can achieve high success rates in the real world, are mainly designed for relatively simple scenes, such as single-object tabletop grasping~\citep{lumdextrah, singh2024dextrah}. These methods, however, aren't directly applicable to cluttered scenes because they rely on simplified object states, like 6D poses, to ensure stable and convergent teacher policy learning, yet fail to capture the essential local geometric details crucial for dexterous grasping in cluttered scenes \citep{dynamic_models, learned_local_models}. To compactly represent the interaction between the object and the dexterous hand, \citep{zhang2024graspxl} proposes a representation that computes the distance between the links of the dexterous hand and the graspable and ungraspable areas, which further improves teacher policy training~\citep{zhang2025robustdexgrasp}.

In this paper, we aim to develop a generalized closed-loop policy for dexterous target-oriented grasping in cluttered scenes, trained in a simulation environment without expert demonstrations \citep{her, deep_rl}, and to achieve zero-shot sim-to-real transfer \citep{domain_randomization, dr, system_identification}. The main challenges are as follows: (1) training instability in RL due to high-dimensional observation and action spaces caused by the diversity of object geometries and layouts, as well as the high DoF of the dexterous hand \citep{deep_rl, deep_rl2, openai_hand}, (2) the sim-to-real gap in observation space, policy safety, and physical dynamics \citep{maddukuri2025sim, lin2025sim, valassakis2020crossing}. To address these challenges, we propose ClutterDexGrasp, a two-stage teacher-student framework. The teacher is trained in simulation using clutter density curriculum learning to reduce the difficulty of directly learning complex grasping strategies. Additionally, we leverage the distance representation~\citep{zhang2024graspxl} and extend it to clutterd scenes by computing distance between links of the dexterous hand and the target and non-target objects, such a geometry and spatially embedded scene representation captures the local geometry of the target object and its surrounding environment, enabling the teacher policy to effectively learn human-like grasping strategies in cluttered scenes \citep{manipulation_behavior}. To ensure safety during contact-rich interactions, we introduce an interaction-aware safety curriculum that minimizes collisions and avoids harmful behaviors, such as excessive force. The student policy is trained using offline imitation learning with a 3D diffusion policy (DP3)~\citep{ze20243d} on partial point cloud observations, enabling zero-shot sim-to-real transfer. 


As shown in Fig.~\ref{fig:teaser}, ClutterDexGrasp achieves general, dynamic, and safe clutter target-oriented dexterous grasping in simulation while also demonstrating zero-shot sim-to-real transfer capability for the first time. 
Our results demonstrate that RL, combined with effective understanding, task representation, safety mechanisms, and efficient distillation \citep{distillation}, can enable robust general dexterous grasping in complex, cluttered environments and facilitate zero-shot sim-to-real transfer.


\section{Related Works}
\label{sec:related_works}

\paragraph{Dexterous Grasping}
Dexterous grasping has progressed from single-object setups to complex cluttered environments \citep{xu2024dexterous,lu2024ugg,zhong2025dexgrasp, general_system_IHM, IHM_survey}. Open-loop methods for cluttered scenes focus on grasp pose generation and planning \citep{wei2022dvgg, zhang2024dexgraspnet, fanganydexgrasp, xu2024dexterous, lu2024ugg}, but lack adaptability. For dynamic control, closed-loop approaches using RL \citep{dexpoint,wu2023learning,wan2023unidexgrasp++,huang2024efficient,lumdextrah,singh2024dextrah, ding2023learning, deep_rl, deep_rl2, openai_hand, openAI_2} achieve success in single-object scenes but face challenges in cluttered environments due to complex interactions and state representations \citep{dynamic_models, learned_local_models, fanganydexgrasp, zhang2024dexgraspnet}. \citep{zhang2024graspxl} and \citep{zhang2025robustdexgrasp} introduce distance between links of the dexterous hand and graspable and ungraspable area of the target object as representation for RL training, with \citep{zhang2025robustdexgrasp} demonstrating great performance on real robots using a teacher-student framework. However, their training is conducted exclusively on single objects, which limits their applicability to complex scenarios. In contrast, our approach directly incorporates cluttered scene representations during RL training, enabling enhanced performance in complex cluttered configurations, particularly under extreme stacking conditions. Alternately, IL-based methods \citep{zhong2025dexgraspvla, il_survey, il_tilde, il_human_hand} require extensive demonstrations that are costly to collect for diverse cluttered scenarios \citep{imitation_demo_difficult, dagger}. Recent advances using vision-language-action frameworks \citep{zhong2025dexgraspvla, black2024pi_0} and motion capture \citep{bunny} aim to reduce demonstration burden but still require significant human effort. Our approach addresses these challenges through RL with specialized representations and curriculum learning that can generalize to cluttered environments without real-world demonstrations.

\paragraph{Sim-to-Real Transfer with Reinforcement Learning}
Transferring RL policies from simulation to real-world remains challenging for dexterous manipulation \citep{openai_hand, valassakis2020crossing, lin2025sim, maddukuri2025sim, yan2025variable}. RL enables learning complex behaviors that are hard to engineer manually \citep{her, abbeel2005exploration, ding2021sim}, but introduces sim-to-real challenges due to distributional shifts in dynamics and observations \citep{deep_rl, deep_rl2}. Techniques like domain randomization \citep{domain_randomization, dr} improve robustness by training across randomized simulation parameters, while system identification \citep{system_identification, huang2024fungrasp} calibrates simulations to better match real-world dynamics. Safety-aware RL is critical for contact-rich behaviors \citep{Tactile_IHM, vff_control, vff_control_region, yan2025variable}, especially in cluttered scenes with unpredictable object interactions. \citep{lin2025sim} and \citep{maddukuri2025sim} address sim-to-real RL for dexterous manipulation, focusing on single-object tasks. Curriculum learning \citep{openAI_2} structures training by gradually increasing task complexity while maintaining robustness and safety. Our method combines these strategies with an interaction-aware safety curriculum that progressively tightens constraints during training for safe real-world deployment.

\paragraph{Diffusion Policy and Teacher-Student Distillation}
Recent advances in generative modeling have introduced diffusion models for policy learning \citep{diffusion_policy, wang2022diffusion, pearce2023imitating, reuss2023goal, ajay2022conditional, janner2022planning, ding2023consistency, yan2025variable}, with significant potential for modeling complex action distributions. \citep{ze20243d} proposed a 3D diffusion policy that operates directly on point cloud observations, making it well-suited for robotic manipulation tasks with partial observations.
Policy distillation \citep{distillation} transfers knowledge from a teacher policy with privileged information to a student policy with limited observation, which is particularly valuable for sim-to-real transfer.  The teacher-student framework \citep{general_system_IHM, lumdextrah}, combined with diffusion policies, enables zero-shot sim-to-real transfer by distilling privileged simulation knowledge into policies that operate with real-world sensory input. Our approach leverages this paradigm, using a privileged teacher policy to generate demonstrations across varying clutter densities, which are then distilled into a point-cloud-based student policy capable of zero-shot transfer to real-world environments.


\section{Problem Formulation}
\label{sec:problem_formulation}
We formulate target-oriented dexterous grasping as an end-to-end learning problem, using RL to train a teacher policy and IL to train a student policy. The learned policies control both a robotic arm and a dexterous hand to grasp target objects in cluttered environments.
For RL, we define a partially observable Markov decision process $(\mathcal{S},\mathcal{O}, \mathcal{A}, R, \mathcal{T}, \mathcal{E}, \gamma)$ with state space $\mathcal{S}$, observation space $\mathcal{O}$, action space $\mathcal{A}$, reward function $R(s,a)$, transition function $\mathcal{T}(s'|s,a)$, and observation function $\mathcal{E}(o|s)$. A stochastic teacher policy $\pi^E(a|o)$ maps observations to actions, optimizing the discounted return $\mathbb{E}_\pi[\sum_{t=0}^\infty \gamma^t r(o_t, a_t)]$. For IL, we learn a student policy $\pi^S$ by minimizing $\mathbb{E}_{(o,a)\sim\mathcal{D}_E}[-\log \pi(a|o)]$ on expert demonstrations $\mathcal{D}_E$ from $\pi^E$. 

\paragraph*{State and Action Spaces}

We consider a tabletop scenario with a 7-DoF robotic arm $\armjoint \in \mathbb{R}^7$ and a 12-DoF dexterous hand $\handjoint \in \mathbb{R}^{12}$. The hand joints consist of 6-DoF actuated finger joints $\joint^{f} \in \mathbb{R}^{6}$, and 6-DoF underactuated finger joints $\joint^{u} \in \mathbb{R}^{6}$. The action space $\aset \subseteq \mathbb{R}^{13}$ encompasses 7D relative changes for the arm joint $\baseaction$ and 6D absolute joint positions for the actuated hand joints $\handaction$ as \citep{huang2023dynamichandoverthrowcatch}.

\paragraph*{Task Simulation}

For each grasping trial, we initially sample a random number of objects from an object prior distribution. The sampled objects are used to construct the cluttered scenes. Then the target object is randomly selected for grasping.

\paragraph*{Observation}
The student observation space $\mathcal{O}=\armjoint \times \joint^{f} \times \obs^{pc}$ consists of the robot joint positions $\armjoint, \joint^{f}$, and a point cloud observation $\obs^{pc} \in \mathbb{R}^{4\times 5120}$, which is generated from the robot's camera (Detailed in Appendix~\ref{supp:sec:student-obs}).
The teacher observation space $\mathcal{O}=\armjoint \times \handjoint \times \obs^E$ with additional privileged observation $\obs^E$, which includes the geometry and spatial representation (Sec.\ref{sec:rep}), along with privileged state information (Detailed in Appendix~\ref{supp:sec:teacher-obs}).


\paragraph*{Objective}

For teacher policy, the RL objective is to find optimal policy $\pi (\policyaction | o), o\sim\mathcal{O}$ that maximizes the expected discounted reward:
\begin{align}
    \pi^{E^*} &= \arg\max_{\pi}
    \E_{\policyaction_t \sim \pi(\cdot | o_t)}
    \left[ \sum_{t=0}^T \gamma^t r(o_t, \policyaction_t)\right] 
    \label{eq:objective} \\
    \pi^{S^*} &= \arg\min_{\pi} \mathbb{E}_{(o,\policyaction)\sim\mathcal{D}^E}[-\log \pi(\policyaction|o)]
    \label{eq:student_objective}
\end{align}
For student $\pi^S$ it optimizes the IL objective with demonstration dataset $\mathcal{D}^E=\{(o, \policyaction)\}, \policyaction\sim \pi^E(\cdot|o) $ sampled by teacher policy.
The above equations pose challenging objectives, since the policy needs to adapt to different objects and layouts.

\begin{figure*}[t!]
\vspace{0.05cm} 
    \centering
    \includegraphics[width=1\linewidth]{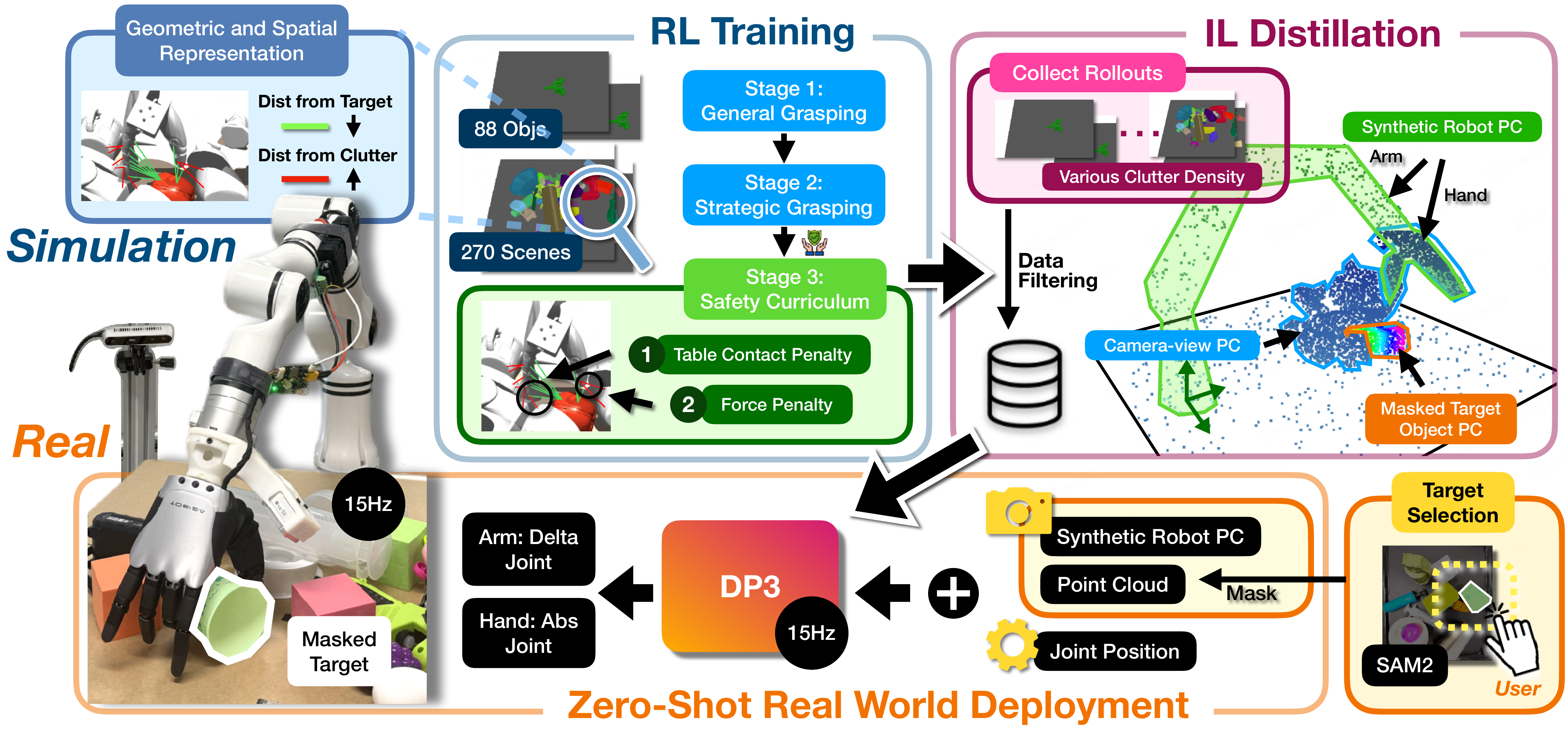}
    \caption{Training Framework}
    \label{Training Framework}
    \vspace{-.1in}
\end{figure*}

\section{Method}
\label{sec:method}

We adopt a teacher-student learning paradigm to develop a policy that is general, dynamic, and safe for dexterous grasping of target objects in cluttered scenes, as illustrated in Fig.\ref{Training Framework}. The framework consists of two main components: (1) a teacher policy trained using privileged state information through our proposed geometry-spatial representation and curriculum learning strategies (\textit{Clutter-Density} and \textit{Safety}), detailed in Sec.\ref{sec:teacher}; and (2) a student policy distilled from successful teacher demonstrations across varying clutter densities. It applies a 3D diffusion policy with point-cloud observations and sim-to-real transfer techniques, described in Sec.\ref{sec:student}. 
The entire training paradigm is conducted in simulation, with no real-world demonstrations used throughout the process.

\subsection{Teacher Policy Learning in Simulation}\label{sec:teacher}

To enable general, dynamic, and safe grasping behavior, we train a teacher policy in simulation using a three-stage curriculum learning strategy. The first two curriculum progressively increases task complexity to guide policy development. The final stage fine-tunes the policy with safety constraints to promote robust and safe execution during real-world deployment. We implement three stages using a unified RL formulation with PPO~\citep{schulman2017proximal} and a novel geometry and spatial representation with privileged state information (Sec.\ref{sec:rep}). Following the objective in Eq.~\eqref{eq:objective}, policies $\pi^E(a|\obs): \obs \rightarrow \Pr(\mathcal{A})$ map privileged observations to actions. We detail each component below.


\subsubsection{Geometry and Spatial Representation}\label{sec:rep}
\paragraph{Cluttered Scene Representation}
For each finger link, we compute the 3D distance vector to the nearest sampled points from both the target ($d_{pos}$) and non-target objects ($d_{neg}$), as illustrated in Fig.~\ref{fig:representation}. This representation efficiently encodes both object geometry and spatial relationships between the hand and the clutter. It enables collision-aware grasping while maintaining a compact observation space that simplifies learning of complex geometric features and avoiding computationally expensive point-cloud rendering. Detailed implementation is provided in Appendix~\ref{supp:subsec:gs-rep}.
\paragraph{Dense Reward Function} 
The reward function with embedded representation at each timestep is:
\begin{equation}
\label{eq:reward_function}
    r = (c_1 \cdot r_\text{grasp} + c_2 \cdot r_{pos}) \cdot (1 - r_{neg})
\end{equation}
where $c_1, c_2 > 0$ are weighting coefficients. Here, $r_{pos}$ provides a reward term encouraging proximity to the target, while $r_{neg}$ imposes a penalty for risky closeness to non-target objects, and $r_\text{grasp}$ represents the base grasping reward (detailed in Appendix~\ref{supp:sec:teacher-reward}).



\subsubsection{Two-Stage Clutter-Density Curriculum}\label{sec:density-curr}

As the first two stages of curriculum learning, we introduce the \textit{Clutter-Density Curriculum}.
Given the complexity of learning general grasping in cluttered scenes, where intricate interactions and robust grasping are required, learning entirely from scratch is found to be infeasible as shown in Fig.~\ref{fig:ablation_without_curriculum}.
To tackle this, we decompose the learning process into two stages: (1) learning robust, generalizable grasping policy for single-object scenarios, and (2) fine-tuning the policy in cluttered environments to learn contact-rich, strategic grasping. This curriculum allows the agent to first efficiently master basic grasping skills for single objects before learning more complex behaviors required for cluttered scenes, such as moving obstacle objects away before grasping the target object. 


\subsubsection{Interaction Safety Curriculum}\label{sec:safety-curr}
As the final stage, we introduce an interaction \textit{Safety Curriculum} to ensure safe and robust grasping in cluttered scenes. 
This curriculum is crucial for sim-to-real transfer, tempering overly aggressive behaviors while preserving human-like strategies. It is designed to minimize collisions and excessive force during contact-rich interactions common in cluttered environments.
Specifically, the policy is fine-tuned by progressively tightening a force threshold $\bar{f}$ whenever the success rate $w$ surpasses a threshold $\bar{w}$. Exceeding this force threshold incurs a sparse penalty term $r_\text{force}$ in the reward and leads to early termination. The reward function Eq.~\eqref{eq:reward_function} is updated with safety penalties as:
\begin{equation}
\label{eq:safe_reward_function}
    r_\text{safe} = r - c_3 \cdot r_\text{force},
\end{equation}
where $c_3 > 0$ are weighting coefficients. To avoid colliding with table, particularly when grasping small objects, we also disabled collisions between the hand and the table in simulation, and terminate the episode if the fingertip penetrates too deeply into the table surface. The implementation details of safety curriculum are deferred to Appendix~\ref{supp:sec:teacher-safe}.


\subsection{Student Point-Cloud Policy Distillation}\label{sec:student}

To enable real-world target-oriented dexterous grasping in cluttered scenes, we distill the teacher policy into a student policy that operates on partial point cloud observations from a single camera. The student policy is trained offline using demonstrations collected across multiple clutter densities (Sec.\ref{sec:multi-level}). To bridge the sim-to-real gap, we employ point cloud alignment for perception (Sec.\ref{sec:pcaug}) and system identification for dynamics (Sec.\ref{sec:si}). 

\subsubsection{Multi-Level Clutter Density Demonstration}\label{sec:multi-level}
The student policy should exhibit adaptive behaviors based on scene clutter density. In sparse scenes, direct grasping is preferred, while in dense scenes, the target-oriented policy must first clear obstacles before grasping. To capture this full spectrum of behaviors, we collect expert demonstrations from the teacher policy $\pi^E$ across varying clutter densities, creating a balanced dataset $\mathcal{D}^E$ for distillation. To model these diverse and complex behaviors, we leverage Diffusion Policy~\cite{diffusion_policy}, which has proven effective for high-dimensional dexterous manipulation~\cite{wu2023graspgf,wu2024unidexfpm}. Specifically, we use DP3~\cite{ze20243d} as our backbone for processing point cloud observations. The student policy $\pi^S$ is optimized by minimizing the negative log likelihood on the expert demonstrations as Eq.~\eqref{eq:student_objective}. The details are deferred to Appendix~\ref{supp:sec:multi}.

\subsubsection{Point Cloud Alignment}\label{sec:pcaug}
To address the sim-to-real perception gap and handle occlusions in cluttered scenes, we augment the observed point cloud with synthetic, dense robot point clouds as shown in Fig.\ref{fig:pointcloud_comparison}. Similar to Dexpoint\citep{dexpoint}, our augmentation includes both arm and hand point clouds, which helps establish critical spatial relationships for accurate grasping and collision avoidance. The details are deferred to Appendix~\ref{supp:sec:pcaug}.

\subsubsection{System Identification for Sim-to-Real Transfer}\label{sec:si}
To minimize the dynamics gap between simulation and reality, we perform system identification (SI)~\citep{system_identification, chen2023visual} to calibrate the physical parameters of both the robotic arm and hand. The SI process iteratively compares trajectories from simulation and the real world under identical commands, 
optimizing parameters until behaviors closely match
~\citep{valassakis2020crossing}, detailed in Appendix~\ref{supp:sec:si}.


\section{Experiments}
\label{sec:result}
\label{sec:experiment}
We conduct comprehensive experiments in both sim and real to validate the following questions:
\vspace{-0.5em}
\begin{itemize} 
    \item Can our framework produce robust and safe policy with high success rates in simulation?
    \item Can our policy be trained in simulation, zero-shot transfer to the real world?
    \item Does the policy generalize across unseen objects and unseen layouts?
 \end{itemize}
 \vspace{-0.5em}

\subsection{Simulation Experiments}

\paragraph{Experiment Setup} The object datasets consist of 88 training objects from GraspNet1Billion~\citep{fang2020graspnet} and 2029 testing objects from Omni6DPose~\citep{zhang2024omni6dpose}. The training set is used to generate 270 training scenes. To evaluate across clutter levels, we create three scene types: sparse ($N_\text{object} \in [4,8]$), dense ($N_\text{object} \in [9,15]$), and ultra-dense ($N_\text{object} \in [16,25]$). For evaluation, we generate 500 unseen layouts using training objects with a sparse:dense:ultra-dense ratio of 3:4:3. For unseen objects, we generate 550 unseen layouts following a 6:3:2 ratio. In each trial, a random visible object is selected as the grasp target. All experiments use three random seeds, and we report the mean and standard deviation of the \textbf{Success Rate}: a trial is successful if the target object is lifted 0.1 meters. Simulation environment details are in Appendix~\ref{supp:sec:sim-env}.

\subsubsection{Main Results}
\begin{table}[!h]
    \begin{center}
        \resizebox{0.9\linewidth}{!}{

\begin{tabular}{cc|c|ccc}
\toprule
\multirow{2}{*}{\textbf{Dataset}} & \multirow{2}{*}{\textbf{Policy}} & \multirow{2}{*}{\textbf{Seen Layouts}} & \multicolumn{3}{c}{\textbf{Unseen Layouts}} \\ \cmidrule(lr){4-6}
& &  & \textbf{Sparse} & \textbf{Dense} & \textbf{Ultra-dense} \\
\midrule
\multirow{2}{*}{\textbf{GraspNet1Billion (Seen)}} & \textbf{Teacher} & $91.9\pm0.3$ & $92.5\pm0.8$ & $87.5\pm0.3$ & $80.9\pm0.2$\\
 & \textbf{Student} & $87.0\pm0.3$ & $89.7\pm0.5$ & $83.4\pm0.3$ & $73.5\pm0.5$\\
\midrule
\multirow{2}{*}{\textbf{Omni6DPose (Unseen)}} & \textbf{Teacher} & / & $92.6\pm0.4$ & $86.6\pm0.4$ & $81.6\pm0.3$\\
& \textbf{Student} & / & $90.8\pm0.7$ & $82.1\pm1.6$ & $74.2\pm1.8$\\
\bottomrule
\end{tabular}


        }
    \end{center}
    \caption{\textbf{Simulation Success Rate of Random Object Grasping}. The unseen clutter scenes are classified into three-density-level: Sparse, Dense, Ultra-dense.}
     \label{table:simulation}
     \vspace{-.3in}
\end{table}

\paragraph{Generalization Across Diverse Clutter Density and Novel Scenes} 
As shown in Tab.~\ref{table:simulation}, our teacher policy achieves a 91.9\% success rate on seen objects and seen layouts. For seen objects with unseen layouts, the teacher policy generalizes well across different clutter densities and achieves comparable results even on unseen objects and unseen layouts, demonstrating the strong generalization capability of our method. The student policy, distilled to partial point cloud observation, shows less than a 5\% average success rate drop across all combinations of seen and unseen objects and layouts, highlighting the effectiveness of our distillation framework. Importantly, both policies were trained only on scenes with sparse and dense clutter levels, yet they generalize well to much harder ultra-dense clutter scenes that were not encountered during training.

\begin{figure*}[h!]
\vspace{-1mm} 
    \centering
    \includegraphics[width=0.8\linewidth]{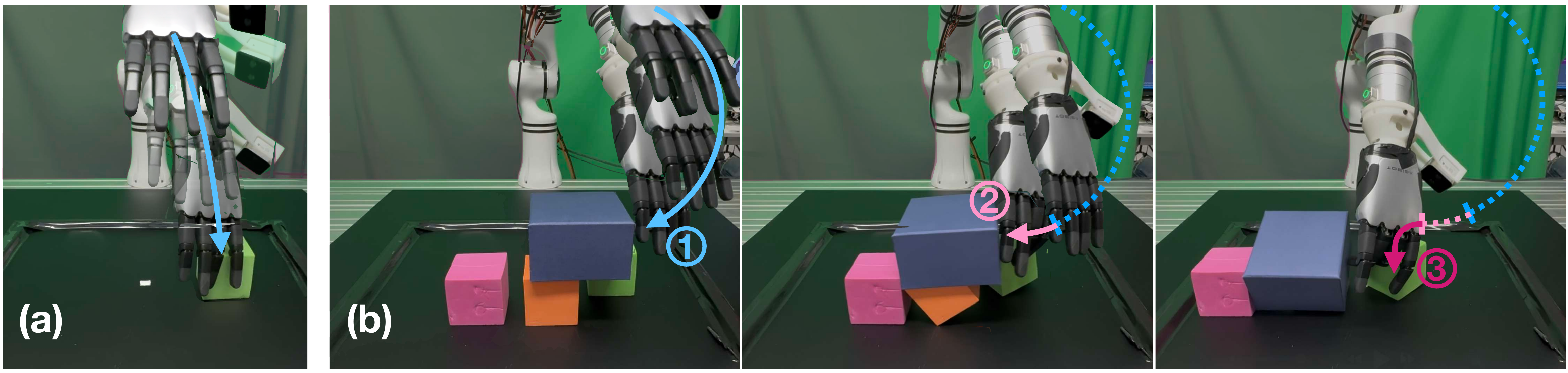}
    \caption{Visualization of Human-like Grasping Strategy: (a) Efficient grasping in simple scenes. (b) Clutter-aware grasping in cluttered scenes.}
    \label{method_trajectory}
    \vspace{-3mm}
\end{figure*}

\paragraph{Human-like Clutter Scene Grasping Strategy}
Quantitative and qualitative results show that the policy learns a human-like grasping strategy for cluttered scenes in both simulation and the real world. Specifically:
(1) \textit{Clutter Clearance via Gentle Interaction}: As shown in Fig.\ref{fig:teaser}, when the target object is deeply buried and direct access is obstructed, the robot gently nudges overlying objects instead of pushing forcefully. As shown in Tab.\ref{table:ablation}, our policy applies low contact force during interaction, further indicating gentle behavior.
(2) \textit{Clutter-aware Grasp}: As shown in Fig.\ref{method_trajectory}, once the obstacles are partially removed, the hand begins to approach from the side to grasp the target object.
Note that these behaviors emerge automatically, without heuristic mode identification or switching, enabling effective, adaptive, and collision-minimized grasping.

\subsubsection{Effectiveness of Density Curriculum and Safety Curriculum}


\begin{wraptable}{tr}{0.4\textwidth}
    \vspace{-5.5mm}
    \begin{center}
        \resizebox{1.0\linewidth}{!}{
            \centering
\resizebox{\columnwidth}{!}{
\begin{tabular}{c|cc}
\toprule
 & \textbf{Success Rate} & \textbf{Force (unit)} \\
\midrule
\textbf{Ours} & $87.0\pm0.3$ & $43.2\pm0.7$ \\
\midrule
\textbf{w/o }\textbf{Safety} & $88.9\pm0.3$ & $80.6 \pm 1.9$ \\
\bottomrule
\end{tabular}
}

        }
    \end{center}
    \vspace{-2.5mm}
    \caption{\textbf{Ablation}: test on unseen layouts with unseen objects.}
    \label{table:ablation_safety}
    \vspace{-4mm}
\end{wraptable}

To evaluate the necessity of the \textit{Density Curriculum}, we conduct an experiment by directly training the policy on cluttered scenes. Under the same training duration, training directly in clutter results in complete failure (0\% success) due to complex interactions, whereas initializing from the general grasp policy achieves 87.0\% success rate—demonstrating the effectiveness of progressive curriculum learning. The learning curve can be found in Fig.~\ref{fig:ablation_without_curriculum}.

To evaluate the \textit{Safety Curriculum}, we measure the average maximum contact forces (in Isaac Gym~\citep{isaac_gym} default force units) across all evaluation trajectories. As shown in Fig.~\ref{fig:ablation_representation_learning_curve}, although the teacher policy without safety training achieves slightly higher success rate (1.9\%), it exhibits excessive force due to risky actions such as poking, jabbing, or squeezing the object, making it unsafe for deployment (Fig.~\ref{fig:policy_performance_vis}). With the safety curriculum, the policy performs more gentle interactions, significantly reduces the force, and shows lower variance (Tab.~\ref{table:ablation_safety}). Crucially, the curriculum filters out overly dynamic, high-risk behaviors—retaining only those that are both human-like and suitable for sim-to-real transfer.

\subsection{Real-world Experiments}

\begin{figure*}[h!]
\vspace{-4mm} 
    \centering
    \includegraphics[width=0.8\linewidth]{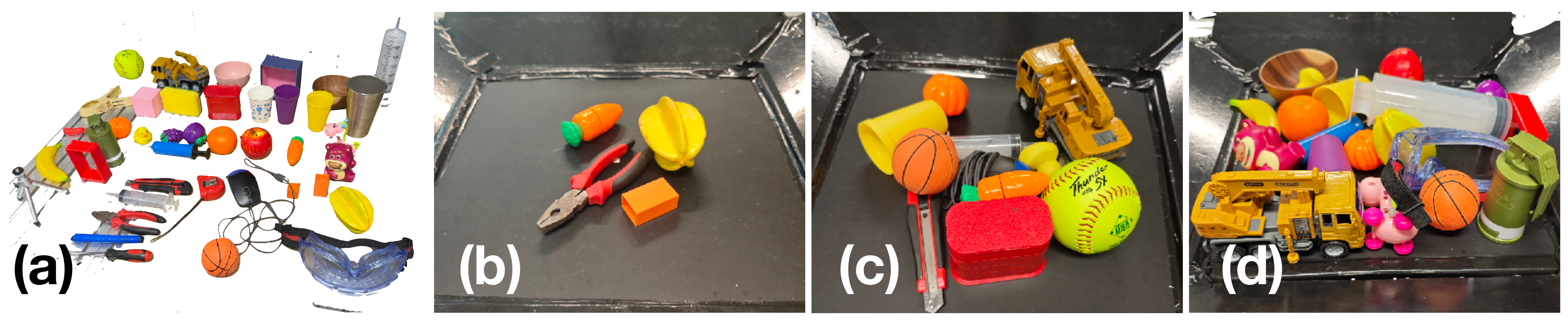}
    \caption{Real-world Objects and Example of Cluttered Scenes: (a): real-world object datasets, (b): sparse cluttered scene, (c) dense cluttered scene, (d) ultra-dense cluttered scene.}
    \label{clutter-scene}
    \vspace{-4mm}
\end{figure*}
\paragraph{Experiment Setup} We chose 41 objects with diverse shapes, sizes, and materials, as shown in Fig.\ref{clutter-scene}. Following the simulation experiments setup, we also conduct grasping under three level clutter density: 9 sparse scenes, 5 dense scenes, and 3 extreme scenes, as shown in Fig.\ref{clutter-scene}. Ensuring that the entire object dataset is covered across all levels of clutter. For each scene, the policy keeps grasping until three consecutive failures happen. For each grasping attempt, the target object is randomly selected from visible masks, increasing difficulty by requiring interaction with occluding clutter. The policy runs at 15 Hz on the real robot. We report \textbf{Success Rate} as $N_\text{successfully grasped objects}/N_\text{total attempts}$ and the \textbf{Area under the Curve (AUC)} to evaluate the efficiency of our system.


\subsubsection{Main Results}

\begin{wrapfigure}{r}{0.42\textwidth}
    \vspace{-3mm}
    \begin{minipage}{\linewidth}
        \centering
        \includegraphics[width=\linewidth]{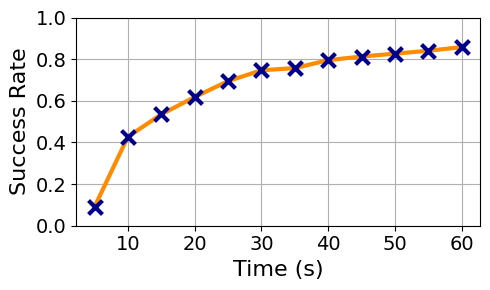}
        \captionsetup{width=0.95\linewidth}
        \caption{Real-world Experiment}
        \label{fig:eff}
        \vspace{1.5mm}
    \end{minipage}
    \begin{minipage}{\linewidth}
        \centering
        \resizebox{\linewidth}{!}{

\begin{tabular}{c|ccc|c}
\toprule
\textbf{Method} & \textbf{SR@20s} & \textbf{SR@40s} & \textbf{SR@60s} & \textbf{AUC} \\
\midrule
\textbf{Ours} & 61.8 & 79.5 & 83.9 & 0.617 \\
\bottomrule
\end{tabular}
        }
        \captionsetup{width=0.95\linewidth}
        \captionof{table}{Quantitative evaluation results in the real world.}
        \label{table:ablation}
    \end{minipage}
    \vspace{-2mm}
\end{wrapfigure}


Our sim-to-real strategy demonstrates strong real-world performance across varying levels of clutter, achieving an overall 83.9\% success rate on unseen layouts with unseen objects over 167 grasping attempts, without any cluttered scene being early-stopped due to three consecutive failures. This performance is comparable to simulation results, confirming the effectiveness of our transfer pipeline. The policy generalizes robustly to a wide range of object geometries and sizes—including irregular and challenging shapes such as a Realsense tripod—without any real-world training. As shown in Fig.\ref{method_trajectory}, the student policy reproduces human-like, contact-aware strategies similar to those seen in simulation. Furthermore, Fig.\ref{fig:eff} illustrates that most of successful grasps are completed within 30 seconds, even in dense clutter, underscoring the system’s efficiency. The cumulative success curve reaches 80\% within 40 seconds, demonstrating that the policy reliably achieves high success rates given sufficient time.





\label{sec:main_results}


\section{Conclusion}
\label{sec:conclusion}
In this paper, we introduce ClutterDexGrasp, a framework designed for dexterous target-oriented grasping in cluttered environments. Utilizing a two-stage teacher-student framework, we train a general, dynamic, and safe teacher policy in simulation with geometry- and spatially-embedded scene representation, via clutter density and safety curriculum, and transfer it to the point cloud-based student policy using a 3D diffusion policy with several sim-to-real strategies. Experimental results demonstrate that the method performs well in simulation and can be zero-shot transferred to real-world environments, effectively generalizing to various objects and layouts for the first time. 


\section{Limitation}
Although our method achieves a high success rate, it still faces limitations with tiny objects due to imprecise grasps caused by the sim-to-real gap and frequent occlusions by the hand and surrounding objects. This highlights the need for enhanced perception strategies—such as multi-view fusion or active vision—to improve visibility and grasp planning in cluttered real-world scenes.

\acknowledgments{We thank Tianyu Wang, Hongwei Fan and Hang Dai for their support in conducting real-world experiments, Hongjie Fang and Tyler Ga Wei Lum for their insightful discussion. We are also grateful to Yanzhou Jin and Haotian Jin for their assistance with the hardware in this project. This research was supported by The National Youth Talent Support Program (8200800081) and National Natural Science Foundation of China (62376006).}


\bibliography{example}  

\newpage
\appendix
\onecolumn
\UseRawInputEncoding
\clearpage
\section{Ablation Study}

\subsection{Geometric and Spatial (GS) Representation} \label{supp:sec:rep-abl}

\begin{figure}[h!]
    \centering
    \includegraphics[width=0.9\linewidth]{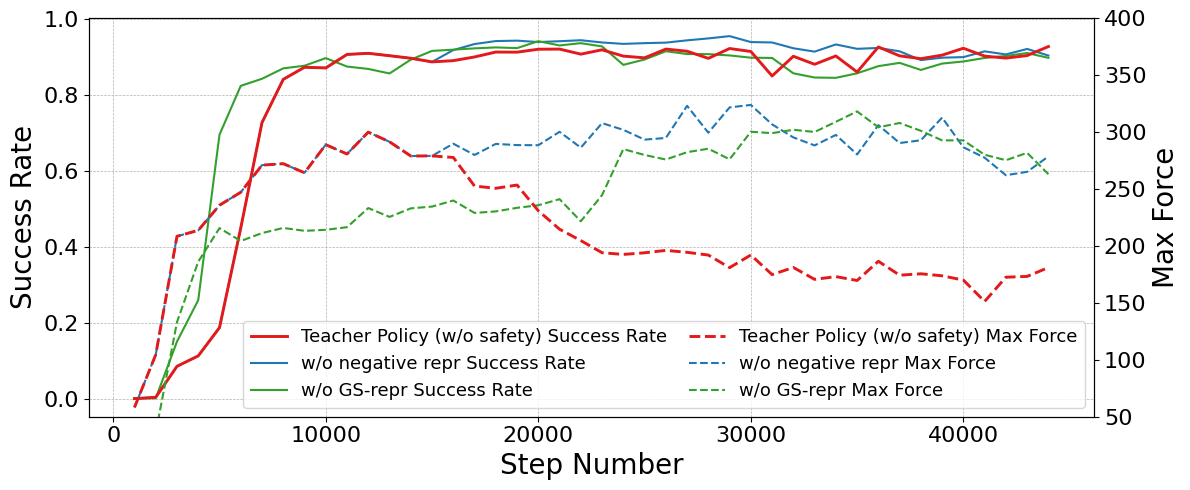}
    \caption{Learning curves of the cluttered-scene policies with or without the Geometric and Spatial Representation we introduced. All experiments shown in this figure were conducted without a safety curriculum.}
    \label{fig:ablation_representation_learning_curve}
\end{figure}

To better understand the contribution of GS-representation, we introduce two ablation experiments with the same framework: (1) one excludes the proposed geometry-and-spatial representation from both observation and reward function (\textit{teacher policy w/o GS-repr}) (2) and another without the distance to non-target object (\textit{teacher policy w/o negative repr}). 

In the \textit{teacher policy w/o GS-repr} setting, we replaced the GS-representation with a simplified distance vector: a distance vector from the object center to the hand palm. The observation still includes the distance to both target and non-target objects but lacks object geometric information. In the \textit{teacher policy w/o negative repr} setting, we use the same geometry-and-spatial representation as in our full method, but mask out features $d_{neg}$ corresponding to surrounding (non-target) objects during cluttered grasping.

Fig.\ref{fig:ablation_representation_learning_curve} shows that while the success rates of these methods are comparable, our method achieves significantly lower maximum contact force as the training progresses, indicating that it learns to avoid collisions with surrounding objects when grasping the target and established safer strategy.
\begin{figure}[h!]
    \centering
    \includegraphics[width=1.0\linewidth]{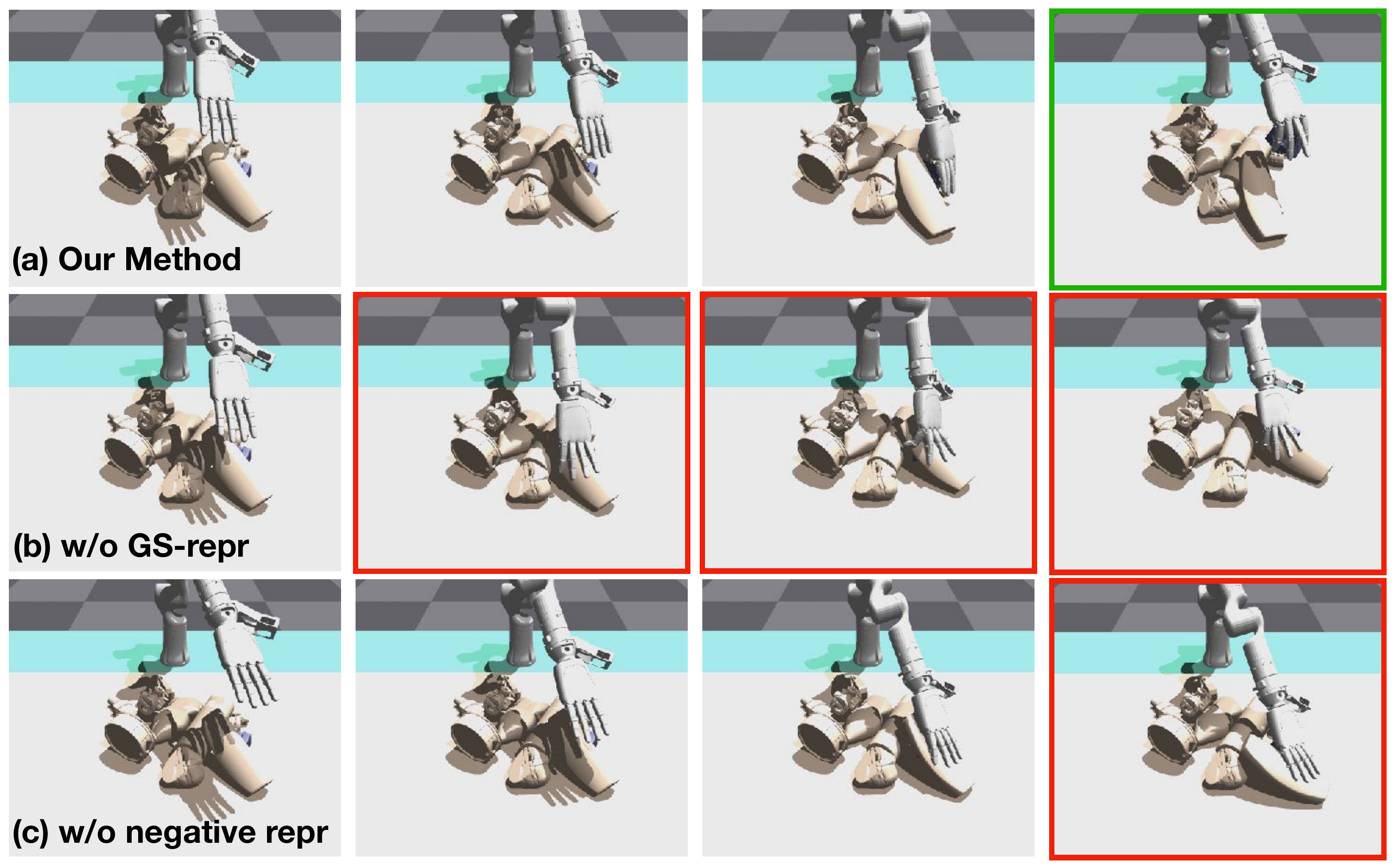}
    \caption{Cluttered Scene Policy Strategy Comparison: (a) Policy trained with GS-Representation, (b) Policy trained without GS-Representation, (c) Policy trained without negative representation. Green bounding boxes indicate successful grasps, while red bounding boxes highlight unsafe or risky actions.}
    \label{fig:policy_performance_vis}
\end{figure}

The qualitative difference is more apparent in rollout visualizations (Fig.\ref{fig:policy_performance_vis}). Both ablated variants, \textit{teacher policy w/o GS-repr} and \textit{teacher policy w/o negative repr}, frequently attempt direct top-down grasps, ignoring clutter and applying excessive force on clutter, especially when the target is partially occluded, leading to unsafe behaviors. In contrast, the GS-based policy exhibits more strategic behavior: gently repositioning occluding objects from the side and approaching the target from angles that minimize collision. Such human-like strategies are critical for sim-to-real transfer. Importantly, this also leads to a smooth transition into the safety curriculum stage, with negligible performance drop (Table\ref{table:ablation_safety}).

\subsection{Clutter-Density Curriculum Learning} \label{supp:sec:cur-abl}
\begin{figure}[h!]
    \centering
    \includegraphics[width=0.9\linewidth]{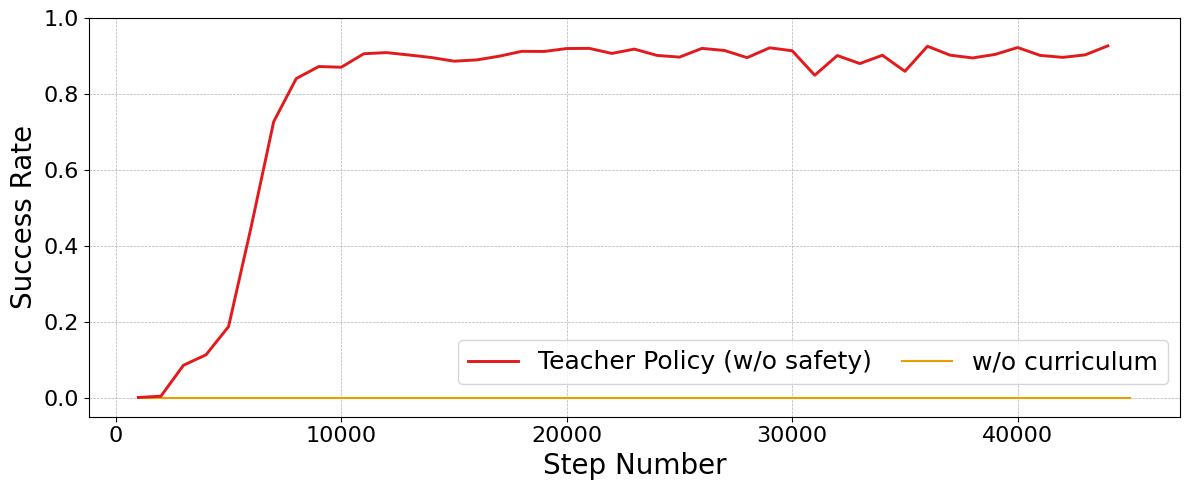}
    \caption{Learning curves of the cluttered-scene policies (1) \textit{Teacher Policy (w/o safety)}: initialized with stage 1 general single-object grasping policy, (2) \textit{w/o curriculum}: trained from scratch directly in cluttered scenes for the full two-stage duration.}
    \label{fig:ablation_without_curriculum}
\end{figure}

Learning dexterous grasping in cluttered scenes purely through random trial-and-error is extremely challenging. As shown in Figure~\ref{fig:ablation_without_curriculum}, a policy trained only on clutter scenes with identical settings and full two-stage duration (\textit{w/o curriculum}) fails to make progress, with success rates remaining at zero. To address this, we introduce a two-stage Clutter-Density Curriculum: the policy is first trained on general single-object grasping, then fine-tuned in cluttered scenes to develop strategic, human-like behaviors. This staged approach enables effective learning, as illustrated by the performance curve in Figure~\ref{fig:ablation_without_curriculum}.

\section{Implentation Details of Teacher Policy Learning} \label{supp:sec:teacher}

\subsection{Geometric and Spatial (GS) Representation}\label{supp:subsec:gs-rep}


Direct grasp attempts in dense cluttered scenes often result in unsafe collisions with surrounding non-target objects, we therefore introduced a Geometric and Spatial Representation to encourage collision-minimized behavior. This representation serves as a \textit{privileged observation} for the \textit{teacher policy}, providing essential spatial and geometric information for clutter-aware grasping, while bypassing the need for time-consuming point-cloud rendering. At each time step, we have $N_{\text{env}}$ environments. For each environment, we sample 200 points from a randomly selected target object, and represent all points as a $N_{\text{env}} \times 200 \times 3$ matrix. We sample 50 points from non-target objects $N_{\text{neg}}$, and represent all points as a $N_{\text{env}} \times 50 \times 3$ matrix. We get 11 base positions from selected finger links, and represent as a $N_{\text{env}} \times 11 \times 3$ matrix. Using torch.norm, we compute distances between link positions and both target object points and non-target object points, identify nearest points, and construct 3D vectors from link bases: $d_{pos}, d_{neg} \in \mathbb{R}^{N_{\text{env}} \times 11 \times 3}$, which serve as observations. During training, $N_{\text{env}} = 16384$ with computation time cost 6ms. This computation is not required by the student policy during real-world deployment.

For single-object grasping scenarios, the representation-related components in both observation and reward are zero-padded for non-target objects.

\begin{figure}[h!]
    \centering
    \includegraphics[width=0.5\linewidth]{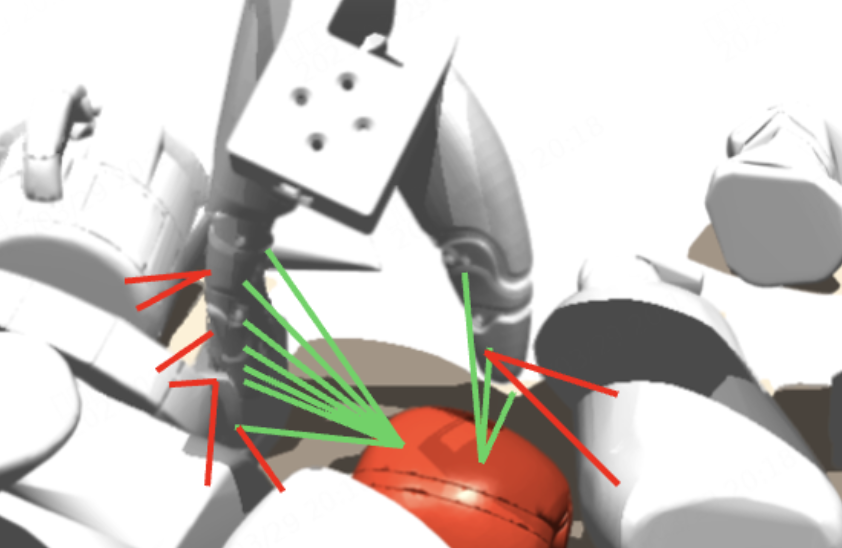}
    \caption{Visualization of the Geometric and Spatial Representation. For each finger joint, distances to the nearest \( N_{\text{pos}} \) surface points sampled from the target object mesh (\( d_{pos} \), green) and \( N_{\text{neg}} \) points from surrounding non-target object meshes (\( d_{neg} \), red) are computed and visualized.}
    \label{fig:representation}
\end{figure}

\subsection{Teacher Observation Space}\label{supp:sec:teacher-obs}
The observation space for teacher $o_t \in\mathcal{O}=\armjoint \times \handjoint \times \obs^E$ is detailed in Tab.~\ref{table:observation_safety}.

\begin{table}[ht!]
    \centering
    \caption{Teacher Observation Space}
    \label{table:observation_safety}
    \begin{tabular}{c|l}
        \toprule
        \textbf{Index} & \textbf{Description} \\
        \midrule
        0 -- 19 & DoF positions (unscaled) $\armjoint \in \mathbb{R}^7$, $\handjoint \in \mathbb{R}^{12}$\\
        19 -- 22 & End-effector position (XYZ) \\
        22 -- 25 & End-effector orientation (Euler angles: roll, pitch, yaw) \\
        25 -- 28 & End-effector linear velocity \\
        28 -- 31 & End-effector angular velocity \\
        31 -- 34 & Vector from object to middle point \\
        34 -- 37 & Middle point position \\
        37 -- 44 & Object pose (position + quaternion) \\
        44 -- 47 & Object linear velocity \\
        47 -- 50 & Object angular velocity \\
        50 -- 57 & Object goal pose (position + quaternion) \\
        57 -- 90 & Distance features (flattened) \\
        90 -- 123 & Negative distance features (flattened) \\
        123 -- 128 & Safety: finger-tip-to-table height (5 values) \\
        \bottomrule
    \end{tabular}
\end{table}

\subsection{Teacher Reward Function} \label{supp:sec:teacher-reward}

We first introduce the definitions of $r_{pos}$ and $r_{neg}$ as:
\begin{equation}
r_{pos} = \exp{(-\alpha_{pos} \cdot \lVert d_{pos} \rVert_2)}
\end{equation}
\begin{equation}
r_{neg} = \exp{(-\alpha_{neg} \cdot \bar{d}_{neg}^{\text{min}})}
\end{equation}
Here, $r_{pos}$ provides a positive reward that encourages the dexterous hand to approach the target object. Conversely, $r_{neg}$ introduces a global penalty term based on $\bar{d}_{neg}^{\text{min}}$, the minimum absolute distance to any non-target (negative) object surface observed throughout the entire grasping episode. This term serves as a global penalty term to discourage risky proximity to non-target objects at any point during execution.
At each timestep, the complete reward function is defined as:
\begin{equation}
\label{eq:reward_function}
    r = \left( c_1 \cdot r_{grasp} + c_2 \cdot r_{pos} \right) \cdot (1 - r_{neg})
\end{equation}
and $r_{grasp}$ is defined as:
\begin{align} 
r_{grasp} &= c_4 \cdot \big( 
    c_5 \cdot (0.2 - \lVert p_{current} - p_{goal} \rVert_2) \tag*{\text{goal distance reward}} \\
    &+ c_6 \cdot \exp{(-\alpha_{mid} \cdot \lVert d_{mid} \rVert_2)} \big) \tag*{\text{middle point reward}} \\
\end{align}
where $c_1, c_2, c_4, c_5, c_6, \alpha_{pos}, \alpha_{neg}, \alpha_{mid} > 0$ are weighting coefficients. $d_{mid}$ is the 3D vector from middle-point between index finger and thumb to center of object. $p_{current}$ and $p_{goal}$ denote the current position of the target object and the desired position the target object should be lifted to after a successful grasp.

\paragraph{Stage 1} For general single-object grasping, the representation-related components in both observation and reward are zero-padded, with reward function being:
\begin{equation}
    r_{stage 1} = c_1 \cdot r_{grasp} + c_2 \cdot r_{pos}
\end{equation}

\paragraph{Stage 2} For strategic cluttered-scene grasping, with reward function is:
\begin{equation}
    r_{stage 2} = r
\end{equation}
where $r$ is defined in Eq.~\ref{eq:reward_function}

\paragraph{Stage 3} For safety finetuning, the reward function for the curriculum is:
\begin{equation}
    r_{stage 3} = r_{safe} = r - c_3 \cdot r_\text{force}
\end{equation}
where \( r_{\text{force}} \) is defined as:
\begin{equation}
    r_{\text{force}} = 
    \begin{cases} 
        1 & \text{if } \max_{i} (f_{z,i}) > f, \\
        0 & \text{otherwise},
    \end{cases}
\end{equation}
with \( f_{z,i} \) representing the z-direction contact force at the \( i \)-th fingertip and \( f \) being the predefined force threshold.

\subsection{Safety Curriculum}\label{supp:sec:teacher-safe}
The safety threshold starts at an initial value of \( f_0 = 200 \) and gradually decreases to a final value of \( \bar{f} = 50 \). After reaching a certain success rate threshold, the safety threshold is reduced by 5 units at each step. To prevent overly rapid updates to the threshold, we define a hyperparameter \( \Delta T_{\text{min}} \), which specifies the minimum number of iterations required between consecutive threshold updates. During the actual training, we disable collisions between the hand and the table, and introduce two early termination conditions: the first is when the contact force of any fingertip exceeds the threshold \( f \), and the second is when any fingertip penetrates the table surface.

\begin{algorithm}
\caption{Safety Curriculum}
\begin{algorithmic}[1]
\State Initialize empty FIFO queue $Q$ of size $K$, $\Delta T = 0$, safety threshold $f = f_0$
\For{$i \gets 1$ to $M$}
    \State $\tau = \text{rollout\_policy}(\pi_\theta)$ \Comment{get rollout trajectory}
    \State $\pi_\theta = \text{optimize\_policy}(\pi_\theta, \tau)$ \Comment{update policy}
    \State $\Delta T = \Delta T + 1$
    \If{$i \bmod L = 0$}
        \State $w = \text{evaluate\_policy}(\pi_\theta)$ \Comment{get success rate $w$}
        \State append $w$ to the queue $Q$
        \If{$\text{avg}(Q) > \bar{w}$ and $\Delta T > \Delta T_{\text{min}}$}
            \State $f = \min(f + \Delta f, f_{\max})$ \Comment{tighten safety constraint}
            \State $\Delta T = 0$
        \EndIf
    \EndIf
\EndFor
\end{algorithmic}
\end{algorithm}

\subsection{Teacher Coarse-to-Fine Data-Efficient Learning}
The high degrees of freedom (DoFs) in dexterous hands pose significant challenges for efficient learning. To address this, we model the grasping process as a coarse-to-fine trajectory: a coarse approach phase followed by a fine interaction phase. During the approach, hand DoFs are frozen by masking hand actions with their initial state, leveraging privileged object-hand distance information $d_{hand}$. Once the hand is sufficiently close to the object $d_{hand}<\bar{d}$, the full action space is enabled. $\bar{d}=0.08$ is used.

\subsection{Training Details}

All the training and experiment in the paper were run on a single GeForce RTX 4090 GPU with i9-13900K CPU. Memory usage and training time are summarized in Table~\ref{tab:computation}, and PPO hyperparameters for the teacher policy are listed in Table~\ref{table:ppo_hyper}.

\begin{table}[h]
    \centering
    \begin{minipage}[t]{0.48\linewidth}
        \vspace{0pt} 
        \centering
        \begin{tabular}{ll}
            \toprule
            \textbf{Hyperparameters} & \textbf{Value} \\
            \midrule
            Num mini-batches & 4096 \\
            Num opt-epochs & 5 \\
            Num horizon-length & 8 \\
            Hidden size & [1024, 512, 256] \\
            Clip range & 0.2 \\
            Max grad norm & 1 \\
            Learning rate & 3e-4 \\
            Discount ($\gamma$) & 0.99 \\
            GAE lambda ($\lambda$) & 0.95 \\
            Init noise std & -- \\
            Desired kl & 0.02 \\
            Ent-coef & 0.0 \\
            \bottomrule
        \end{tabular}
        \vspace{2mm}
        \caption{Hyperparameters of PPO.}
        \label{table:ppo_hyper}

        \vspace{4mm}
        \begin{tabular}{@{}c|c|c@{}}
            \toprule
             & Mem. & Time \\
            \midrule
            Teacher & 22G & 8 days \\
            Student & 22G & 1 day \\
            \bottomrule
        \end{tabular}
        \vspace{2mm}
        \caption{Computation resources.}
        \label{tab:computation}
    \end{minipage}
    \hfill
    \begin{minipage}[t]{0.48\linewidth}
        \vspace{0pt} 
        \centering
        \begin{tabular}{ll}
            \toprule
            \textbf{Hyperparameters} & \textbf{Value} \\
            \midrule
            Downsample dims & [128, 256, 384] \\
            Encoder output dim & 64 \\
            Crop shape & [80, 80] \\
            Horizon & 4 \\
            Num observation steps & 2 \\
            Num action steps & 1 \\
            Kernel size & 5 \\
            Num groups & 8 \\
            Diffusion steps (training) & 100 \\
            Diffusion steps (inference) & 10 \\
            Learning rate & 1e-4 \\
            Optimizer & AdamW \\
            Weight decay & 1e-6 \\
            Betas & (0.95, 0.999) \\
            EMA power & 0.75 \\
            EMA max value & 0.9999 \\
            LR scheduler & cosine \\
            LR warmup steps & 500 \\
            \bottomrule
        \end{tabular}
        \vspace{2mm}
        \caption{Key Hyperparameters of Simple DP3 Policy.}
        \label{table:dp3_hyper}
    \end{minipage}
\end{table}

\section{Implementation Details of Student Policy Distillation}

We offline distilled the teacher policy $\pi^E$ trained with privilage state information $\mathcal{O}^E$ into a student policy $\pi^S$ that takes sensory observations $\mathcal{O}^S \in \mathbb{R}^{4109} $ that can be obtained in the real world. The student observation space contains the robot joint position $\mathcal{O}_{robot} \in \mathbb{R}^{13}$ and the partial point cloud $\mathcal{O}_{pc} \in \mathbb{R}^{4096}$ from a fixed side-view camera cropped and transformed to the robot frame. 

\subsection{Student Observation} \label{supp:sec:student-obs}
The point cloud observation $\obs^{pc}=\obs^p\times \obs^g\times \obs^s$ is composed of three components: (1) $\obs^p \in \mathbb{R}^{4\times 3584}$, the partial point cloud observation from a single third-person camera, with a 1D mask to indicate the grasping target; (2) $\obs^g \in \mathbb{R}^{4 \times 512}$, a synthetic ground point cloud replacing the table surface to mitigate sensor noise; and (3) $\obs^s \in \mathbb{R}^{4\times 1024}$, the synthetic robot point cloud observation (Sec.\ref{sec:pcaug}) with a 1D mask to differentiate between real and synthetic point clouds. The visualization of the point-cloud can be found in Fig.~\ref{Training Framework}.

\subsection{Multi-Level Clutter Density Dataset Generation} \label{supp:sec:multi}
We collect a dataset $\mathcal{D}^E$ consisting of 20,000 successful trajectories generated by the teacher policy $\pi^E$ across 500 clutter scenes with varying levels of clutter density. Demonstrations are evenly distributed across the different clutter levels. The student policy is trained using Imitation Learning (IL) with the DP3 algorithm\citep{ze20243d}, where a random batch size of 120 is sampled from the dataset. Observations are normalized based on the data distribution's statistics. Detailed training hyperparameters can be found in Tab.~\ref{table:dp3_hyper}

\subsection{Point Cloud Observation Processing} \label{supp:sec:pcaug}
The point cloud is created from a single-view depth image. The point-cloud pre-processing includes four steps: (i) cropping the point cloud to the workspace region using a manually defined bounding box, (ii) downsampling the point cloud to 3584 points, (iii) transforming the point cloud from the camera frame to the robot base frame, and (iv) replacing the table surface with a synthetic point cloud (512 points) to mitigate point cloud holes caused by the flat table. We apply consistent point cloud preprocessing across simulation and real-world data to ensure alignment. However, we found that adding Gaussian noise and random transformation to point cloud during DP3 training did not improve policy performance in real-world settings.

\begin{figure}[h]
    \centering
    \begin{subfigure}[t]{0.48\linewidth}
        \centering
        \includegraphics[width=\linewidth]{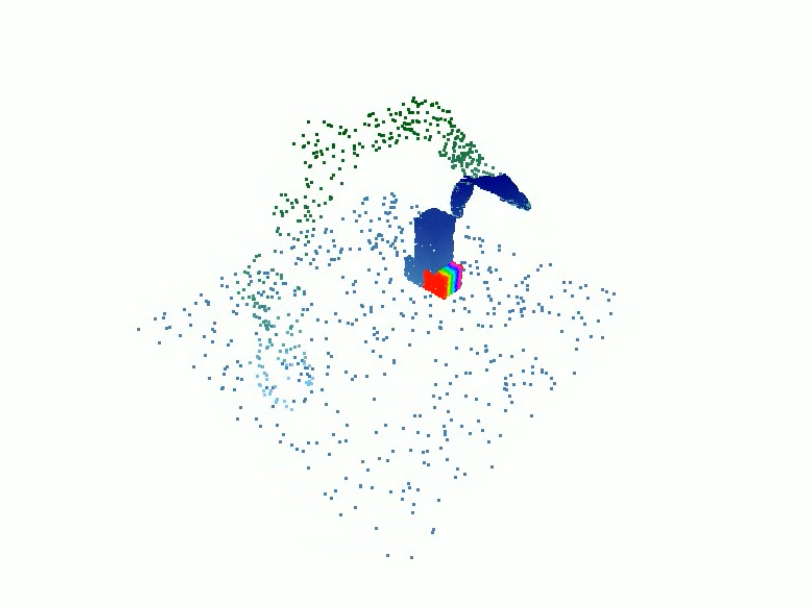}
        \label{fig:sim}
    \end{subfigure}
    \hfill
    \begin{subfigure}[t]{0.48\linewidth}
        \centering
        \includegraphics[width=\linewidth]{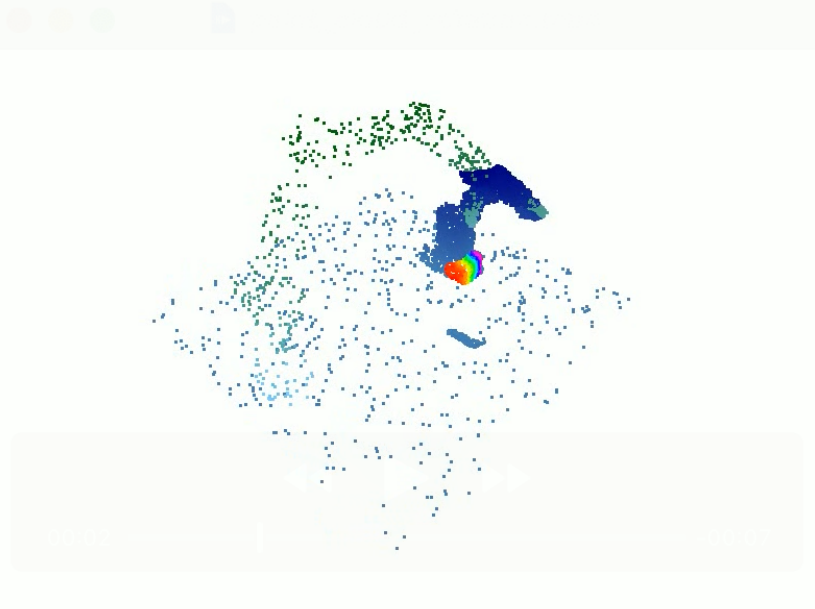}
        \label{fig:real}
    \end{subfigure}
    \caption{Point-cloud comparison between simulation(Left) and real world(Right).}
    \label{fig:pointcloud_comparison}
\end{figure}

\subsection{System Identification}
\label{supp:sec:si}
To minimize the sim-to-real gap, we calibrate the physical and dynamic parameters of robot via system identification (SI) by aligning simulated trajectories with their real-world counterparts, following the approach in \cite{chen2023visual, yan2025variable}.

\subsection{Training Details}
Memory usage and training time are summarized in Table 6, and the hyperparameters for student policy training are listed in Tab.~\ref{table:dp3_hyper}.

\section{Simulation Environment Generation Details}
\label{supp:sec:sim-env}

To generate cluttered tabletop scenes for training and evaluation, we follow a physics-based simulation pipeline. Specifically, we sequentially drop a set of randomly selected objects from a fixed height above the workspace. After all objects are dropped, we allow the simulation to run until the scene reaches a physically stable state. Only those scenes where all objects remain on the tabletop surface are retained for further use. Our simulated robot setup consists of a RealMan RM75-6F 7-DoF robotic arm, equipped with the AgiBot 6-DoF dexterous hand.

\section{Implementation Details in Real-World}
\subsection{Real-World Hardware Setup}

For real-world deployment, the target object is selected and segmented using SAM2~\cite{ravi2024sam2segmentimages}. The resulting binary mask is projected onto the point cloud using one-hot encoding to isolate the target object. 
The entire system operates at a frequency of 15 Hz on a single GeForce RTX 4090 GPU with i9-13900K CPU.
\begin{figure}[h!]
    \centering
    \includegraphics[width=0.7\linewidth]{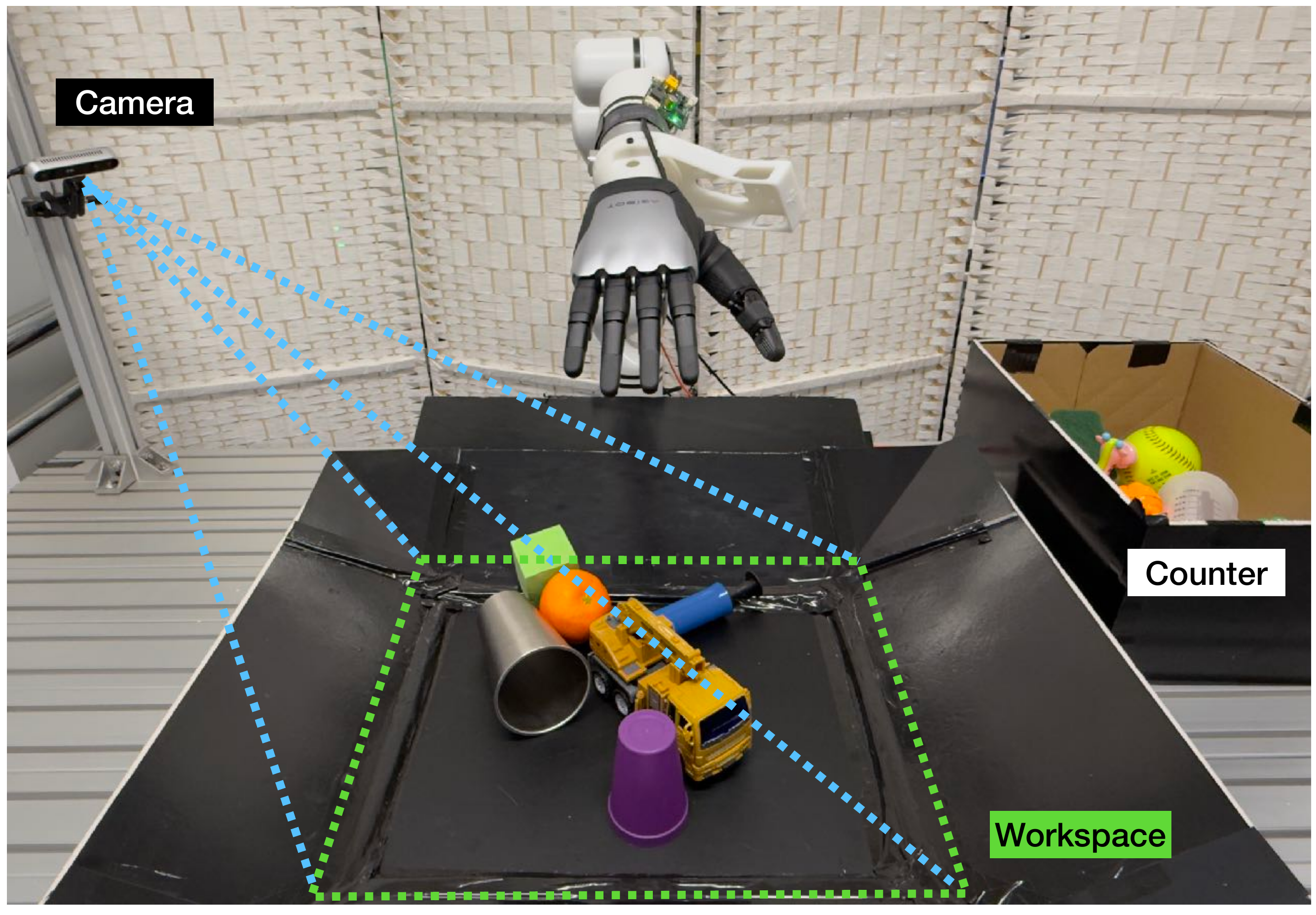}
    \caption{Real-World Setup}
    \label{fig:real_world_setup}
\end{figure}


To improve deployment stability, we introduce several practical adjustments. Since the system occasionally failed to detect successful grasps, leading to redundant actions, we use finger torque as a success signal: Once it exceeds a threshold, the grasp is considered successful and a predefined lift-up action is executed. To guarantee safety, we add a threshold on end-effector force; if exceeded, the end-effector is gently raised to avoid approaching dangerous force boundaries. Additionally, we slow down the end-effector slightly compared to simulation to ensure smoother and safer real-world execution. Note that, in all real-world experiments, no safety-related human intervention was observed.

\section{Failure Case Analysis}

Our policy struggles with grasping extreme shapes and sizes due to the limitations of hand morphology. Specifically, the policy fails with large objects or excessively flat ones. Additionally, due to the limited working space of our robotic arm, we did not strictly constrain the scene generation to the arm's operational range. As a result, some failures occur when objects fall outside the working area or are pushed out of the range by the arm during grasping. Check our website for failure videos.

\begin{figure}[h!]
    \centering
    \includegraphics[width=0.7\linewidth]{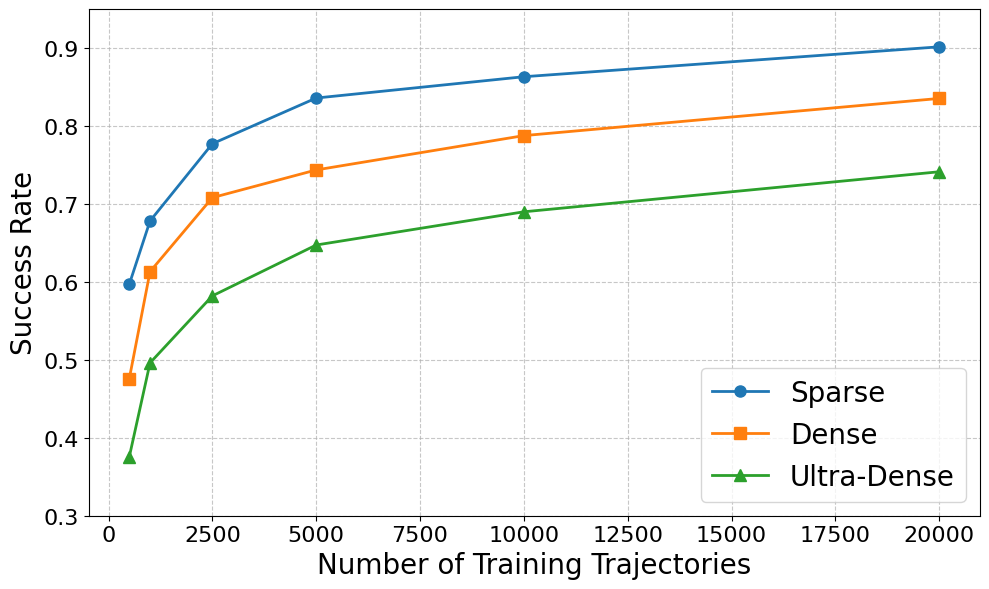}
    \caption{Data Scaling for Generalization}
    \label{fig:student_scaling}
\end{figure}

\section{Data Scaling for Scene-Level Generalization}
\label{supp:sec:data_scaling}
In this section, we investigate how the performance of student policy scales with the number of collected trajectories. Specifically, we use a consistent teacher policy to collect trajectories in identical scenes, but employ varying scales of trajectory data for student policy distillation. Consistent with the main manuscript, we conduct evaluations across three scenario types: Sparse, Dense, and Ultra-Dense. As illustrated in Figure~\ref{fig:student_scaling}, results from these representative scenes consistently show significant performance improvements as the data scale increases. This enhancement in performance is attributed to the richer diversity and more comprehensive coverage of the state-action space afforded by larger datasets, which facilitates more effective learning and generalization.


\end{document}